\documentclass[letterpaper]{article} 
\usepackage{aaai2026}  
\usepackage{times}  
\usepackage{helvet}  
\usepackage{courier}  
\usepackage[hyphens]{url}  
\usepackage{graphicx} 
\usepackage{natbib}  
\usepackage{caption} 
\usepackage{algorithm}
\usepackage{algorithmic}
\usepackage{booktabs}
\usepackage{multirow}
\usepackage[table]{xcolor}
\usepackage{tcolorbox}

\usepackage{newfloat}
\usepackage{listings}
\DeclareCaptionStyle{ruled}{labelfont=normalfont,labelsep=colon,strut=off} 
\floatstyle{ruled}
\newfloat{listing}{tb}{lst}{}
\floatname{listing}{Listing}
\title{CoLM: Collaborative Large Models via A Client-Server Paradigm}
\author{
    Siqi Huang\textsuperscript{\rm 1,2},
    Sida Huang\textsuperscript{\rm 1,2},
    Hongyuan Zhang\textsuperscript{\rm 1,3}
    \footnote{: Corresponding authors. }
}
\affiliations{
    \textsuperscript{\rm 1}Institute of Artificial Intelligence (TeleAI), China Telecom\\
    \textsuperscript{\rm 2} School of Artificial Intelligence, OPtics and ElectroNics (iOPEN), Northwestern Polytechnical University\\
    \textsuperscript{\rm 3}The University of Hong Kong\\
    \{4777huang, sidahuang2001, hyzhang98\}@gmail.com
}

\usepackage{bibentry}

\begin{document}

\maketitle

\begin{abstract}
    Large models have achieved remarkable performance across a range of reasoning and understanding tasks. 
    Prior work often utilizes model ensembles or multi-agent systems to collaboratively generate responses, effectively operating in a server-to-server paradigm. 
    However, such approaches do not align well with practical deployment settings, where a limited number of server-side models are shared by many clients under modern internet architectures.
    In this paper, we introduce \textbf{CoLM} (\textbf{Co}llaboration in \textbf{L}arge-\textbf{M}odels), a novel framework for collaborative reasoning that redefines cooperation among large models from a client-server perspective. 
    Unlike traditional ensemble methods that rely on simultaneous inference from multiple models to produce a single output, CoLM allows the outputs of multiple  models to be aggregated or shared, enabling each client model to independently refine and update its own generation based on these high-quality outputs.
    This design enables collaborative benefits by fully leveraging both client-side and shared server-side models.
    We further extend CoLM to vision-language models (VLMs), demonstrating its applicability beyond language tasks. 
    Experimental results across multiple benchmarks show that CoLM consistently improves model performance on previously failed queries, highlighting the effectiveness of collaborative guidance in enhancing single-model capabilities.
\end{abstract}


\section{Introduction}

\begin{figure}[t]
    \centering
    \includegraphics[width=0.9\linewidth]{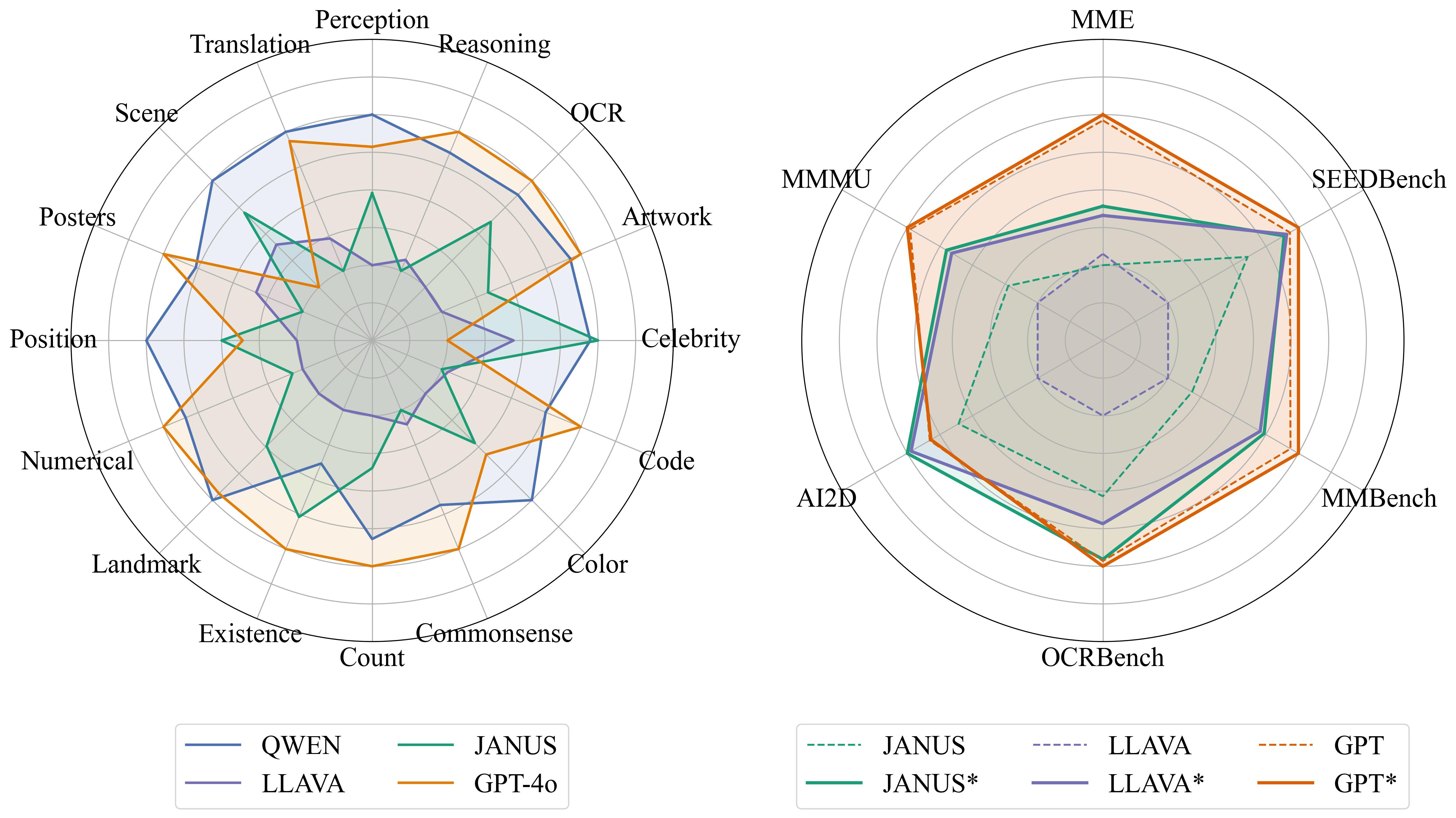}
    \caption{(a) Initial model performance across MME shows no single model excels universally. (b) Models enhanced by our method (marked with *) show consistent improvements across datasets.}
    \label{fig:Radar}
\end{figure}

Large language models (LLMs)~\citep{gpt3, gpt4, deepseek, qwen} and vision-language models (VLMs)~\citep{Gpt-4o, blip2, Qwen2.5-VL} have demonstrated impressive capabilities across a wide range of tasks, including language understanding, logical reasoning, code generation, and multimodal question answering. 
Their performance often improves with scale, making them a foundation for many state-of-the-art systems in artificial intelligence.
However, growing evidence from recent empirical studies and deployment experiences suggests that no single model can consistently dominate across all domains, task types, or input distributions~\citep{singleno1, singleno2, singleno3}, as illustrated in our Figure~\ref{fig:Radar}(a).
Even state-of-the-art models may struggle with out-of-distribution inputs or domain-specific requirements. 
To further enhance the reasoning ability of large models, prior work has explored collaborative approaches, such as model ensembles, which aggregate outputs from multiple models, and multi-agent frameworks, where multiple specialized or redundant agents communicate to solve a problem jointly~\citep{moa, selfmoa, think}. 
These methods have shown promising results in boosting accuracy, robustness, and coverage, particularly in tasks where a single model might fail due to uncertainty or lack of context.

However, most existing collaborative paradigms rely on a server-to-server collaboration assumption, where multiple large models can communicate freely and synchronously during inference. 
While these methods are effective in controlled or offline environments, it becomes impractical for large-scale deployment of LLMs and VLMs over the internet, where computational resources are limited and user access is typically routed through shared servers. 
These practical constraints call for a fundamental reevaluation of how collaborative reasoning should be designed in real-world systems. 
In many deployment scenarios such as mobile applications, edge devices, or shared computing clusters, a vast number of clients must interact with only a small number of centralized, high-capacity server models. 
Under such conditions, these methods are incompatible with the nature of client-server architectures, limiting their practicality.

To address these challenges, we propose \textbf{CoLM} (\textbf{Co}llaboration in \textbf{L}arge-\textbf{M}odels), a novel framework that redefines collaborative reasoning from a client-server perspective. 
Unlike traditional ensemble methods that directly produce joint outputs, CoLM leverages intermediate outputs generated by multiple models running on distributed clients, which are then aggregated and refined by a more capable server-side model. 
The resulting guidance is sent back to the client, where  typically lightweight or privacy-constrained target models utilize this information to generate the final responses. 
By separating heavy reasoning from local generation, CoLM enables efficient collaboration that improves client-side performance without incurring the computational cost of ensemble inference or requiring full server-side decoding.

For language tasks, we adopt a three-stage client-server paradigm. 
Domain-specialized client models generate reference responses independently, which are then synthesized by a central server model to produce a global answer. 
This answer is subsequently returned to the clients as guidance, enabling them to refine their final responses. 

Vision-language models are often trained on diverse multimodal datasets and exhibit varied strengths and biases across tasks. 
Rather than forcing ensemble decoding, CoLM for VLMs uses a prompt-based collaboration strategy: the outputs from multiple VLMs are concatenated as contextual input to guide a model. This design naturally supports task-level diversity and allows each model to contribute complementary perspectives, resulting in more robust and accurate final outputs.
Our experiments demonstrate that models with complementary strengths can collaborate to guide a target vision-language model, significantly improving accuracy especially on challenging queries where standalone models often fail.
This demonstrates the potential of client-server collaboration to advance the capabilities and applicability of large models.

\section{Related Work}

\paragraph{Ensemble and Collaborative Reasoning in Large Models}
LLMs have made significant progress in reasoning tasks with prompting techniques such as Chain-of-Thought (CoT)\citep{cot1, cot2} and Self-Consistency\citep{selfconsistency}, which promote step-by-step thinking and improve answer reliability through path sampling.
In parallel, recent studies have explored multi-agent collaboration to enhance LLM reasoning. A common direction involves debate-style frameworks, where multiple models interact through iterative discussion or voting to arrive at better answers~\citep{debate1, debate2}. 
Several works have shown that introducing constructive noise into the model inputs or intermediate representations can enhance model robustness and generalization~\citep{pinoise, vpn, pinda, PiNGDA, MiN}. 
Other approaches, such as Multi-Agent (MoA) and Self-MoA~\citep{moa, selfmoa}, improve prediction by aggregating responses from multiple rounds of model interaction. While these systems rely on iterative collaboration, they do not aim to improve or guide a specific target model.

\paragraph{Routing and Cascading Inference for Cost Efficiency}
To reduce the cost of LLM deployment, routing and cascading methods have been widely explored. Routing methods like RouteLLM~\citep{Routellm} and Eagle~\citep{zhao2024eagle} aim to dynamically select the most appropriate model per input query.
Cascading methods instead involve sequential invocation of models based on response confidence or quality thresholds. FrugalGPT~\citep{chen2024frugalgpt}, for example, uses a judging model to determine if the current model's output is sufficient, invoking stronger models only when needed. Other works explore policy learning~\citep{zhang2024efficient} or structured representations~\citep{yue2024large} to optimize cascade decisions. While effective in saving cost, these approaches still incur multiple inference calls and primarily focus on model switching, not on improving a given model’s capability.
\begin{figure*}[t]
    \centering
    \includegraphics[width=\textwidth]{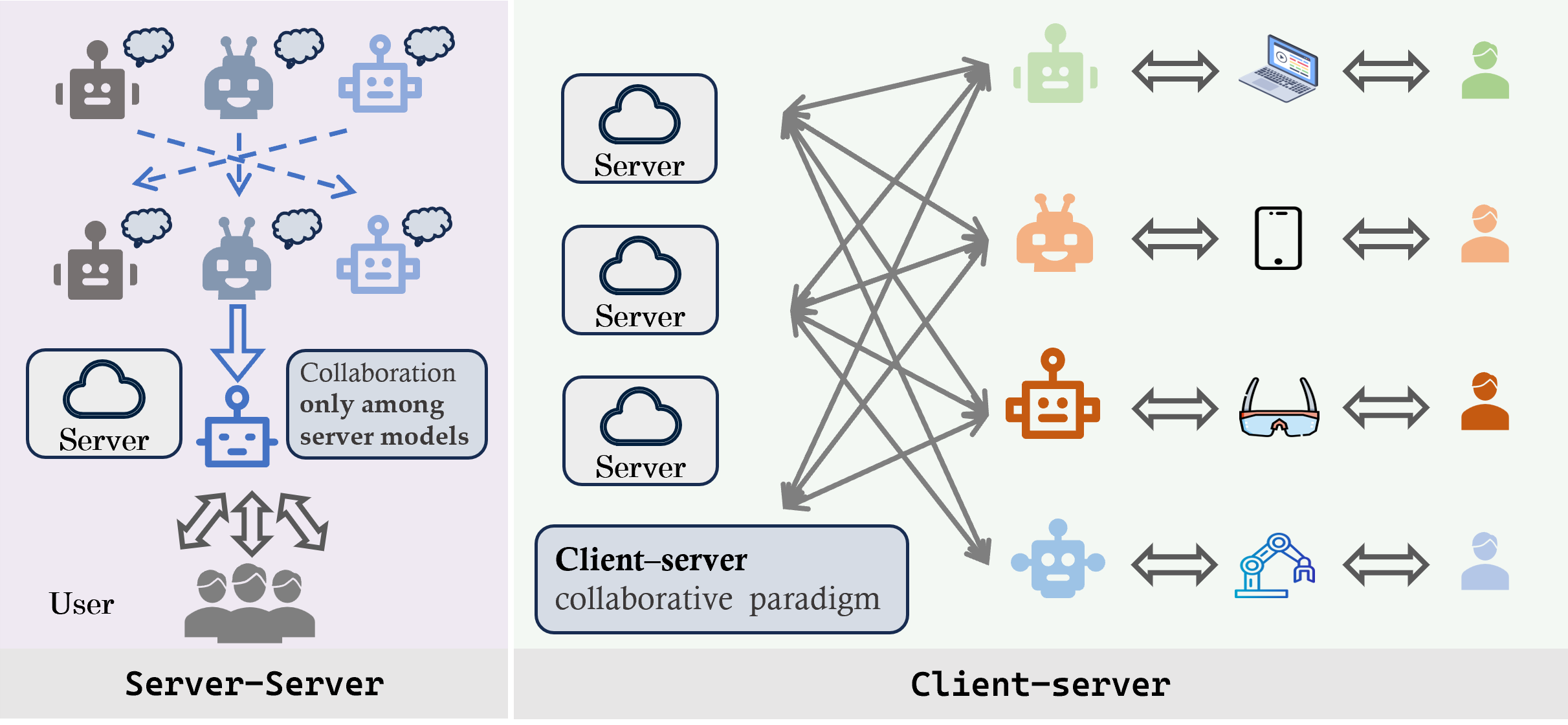}
    \caption{\textbf{Left:} Traditional server-to-server collaboration paradigm, where multiple large models interact directly during inference. These approaches often rely on interactions among general-purpose models, lacking specialization structure.
    \textbf{Right:} Our proposed client-server collaboration paradigm, where lightweight client models receive guidance from shared server-side models. 
    This design allows each client to maintain long-lived, domain-specific expertise while improving response quality through collaboration.}

    \label{fig:FW}
\end{figure*}

\paragraph{Client-Server and Distributed Inference}
Some efforts have explored distributed inference strategies to balance latency and computation in real-world systems. Neurosurgeon~\citep{kang2017neurosurgeon} and DDNN~\citep{teerapittayanon2017distributed} propose splitting models between edge and cloud for collaborative computation. Auto-split~\citep{Auto-split} further generalizes this concept into a practical framework for edge-cloud AI deployments. 
A comprehensive survey~\citep{wang2024end} reviews recent advances in end-edge-cloud collaborative deep learning, highlighting challenges and system-level design considerations. These approaches align with our CoLM design, which leverages limited server-side interaction to guide lightweight client-side inference, enabling scalable, efficient collaboration in practical deployments.

\paragraph{Collaborative Reasoning in Multimodal Settings}
Some works have explored using multiple agents with distinct capabilities to tackle complex multi-modal tasks. 
Multi-Agent VQA \citep{jiang2024multiagentvqaexploringmultiagent} employs a cooperative setup where a central vision-language model offloads subtasks like object detection or counting to specialized models. 
Similarly, BuboGPT \citep{zhao2023bubogptenablingvisualgrounding} integrates an off-the-shelf grounding module into a multimodal LLM to enhance fine-grained object localization during response generation. 
MMCTAgent \citep{kumar2024mmctagent} further advances this approach by introducing a critic module and iterative reasoning loops, mimicking human critical thinking to refine complex visual answers. 
Recent works demonstrate that introducing beneficial noise into multimodal representations can improve alignment and generation quality~\citep{Enhance, Nfig, Adv-cpg, fu2025object}. 
More recently, MAMMQA \citep{rajput2025rethinking} generalizes multi-agent collaboration to handle text, tables, and images jointly, where dedicated VLM and LLM agents sequentially decompose, synthesize, and integrate modality-specific insights.

\section{Method}
\label{sec:method}

Our proposed \textbf{CoLM}  is designed to facilitate efficient and structured collaboration between models deployed in a client-server architecture.
This design closely aligns with practical deployment scenarios where users interact with AI models running on resource-constrained client devices such as smartphones, tablets, or edge computing nodes.

\subsection{Motivation}
In real-world deployment scenarios, client-side models are highly customized, either through domain-specific or long-term adaptation to particular user behaviors. 
These models run on personal devices or localized environments and accumulate specialized knowledge and exhibit strong domain preferences.
They distinct from one another not just in scale, but also in perspective and reasoning habits.

Unlike traditional ensemble or agent systems which typically rely on multiple models to reach a consensus, our setting embraces the diversity among client models. 
This diversity is not a source of noise but rather a valuable feature. Each model reflects a unique domain expertise or personalization history. 
In our setup, client models first independently produce responses based on their specialized understanding. These responses are then sent to a central server model, which integrates them to generate guidance. Then each client model can use to revise and refine its own answer.

This interaction loop encourages models to not only contribute their strengths but also evolve through exposure to alternative perspectives. It allows underperforming models to benefit from others' knowledge, while still preserving their personalized traits. In doing so, CoLM enables a richer form of collaboration that improves robustness and generalization.
As shown in our experiments (Section~\ref{sec:experiments}), CoLM achieves a strong balance between collaboration efficiency and performance, demonstrating its practicality and broad potential for real-world deployment.

\subsection{The CoLM Inference Pipeline}
CoLM supports collaborative reasoning in both language and vision-language scenarios. While both share the same guiding principle, the actual inference pipelines differ due to architectural differences between LLMs and VLMs.

For LLM models, CoLM organizes models into two asymmetric roles: lightweight client models that generate reference responses, and a central server model responsible for synthesizing and producing the final output.

Given a user query $q$, the inference process begins by identifying which models in a larger pool are most relevant to the query. Let $\mathcal{C} = \{ M_1, M_2, \dots, M_K \}$ denote the complete set of available client models. Each client $M_i$ is associated with a specialization prompt $\mathcal{P}(M_i)$ that describes its intended domain or behavioral role (e.g., “You are an expert in math”). We use a strong general-purpose language model (e.g., GPT-4o) to estimate the semantic similarity between $q$ and each $\mathcal{P}(M_i)$, and select the top-$k$ most relevant models to form a task-specific subset $\mathcal{C}^* \subseteq \mathcal{C}$.

\begin{table*}[t]
\centering
\renewcommand{\arraystretch}{1.3}
\resizebox{\textwidth}{!}{
\begin{tabular}{l||c|c|c|c|c|c|c|c|c}
\hline\toprule
\textbf{Model} & \textbf{MME-P} & \textbf{MME-R} & \textbf{SEEDBench} & \textbf{MMBench} & \textbf{OCRBench} & \textbf{AI2D} & \textbf{MMMU-Val} & \textbf{MMMU-Dev}& \textbf{Avg. Score} \\
\midrule
Qwen2.5-VL-7B      & \textbf{1693.53} & 611.43 & 0.771 & \textbf{0.831} & \textbf{881} & 0.809 & 0.444 & 0.433 & 61.88\\
\rowcolor{gray!10}
Qwen2.5-VL-7B*     & 1656.04 & \textbf{614.64}↑ & \textbf{0.772}↑ & 0.819 & 865 & 0.782 & \textbf{0.532}↑ & \textbf{0.513}↑ & \textbf{63.30}↑\\

Janus-Pro-7B       & \textbf{1509.38} & 270.71 & 0.701 & 0.665 & 584 & 0.679 & 0.380 & 0.373 & 47.55\\
\rowcolor{gray!10}
Janus-Pro-7B*      & 1482.28 & \textbf{434.64}↑ & \textbf{0.747}↑ & \textbf{0.773}↑ & \textbf{800}↑ & \textbf{0.782}↑ & \textbf{0.499}↑ & \textbf{0.460}↑ & \textbf{58.06}↑\\

LLaVA-1.5-7B       & 1340.31 & 302.14 & 0.601 & 0.629 & 308 & 0.519 & 0.323 & 0.273 & 38.62\\
\rowcolor{gray!10}
LLaVA-1.5-7B*      & \textbf{1349.79}↑ & \textbf{408.93}↑ & \textbf{0.751}↑ & \textbf{0.767}↑ & \textbf{678}↑ & \textbf{0.775}↑ & \textbf{0.489}↑ & \textbf{0.487}↑  & \textbf{56.13}↑\\

GPT-4o             & 1618.96 & 672.86 & 0.755 & 0.813 & 806 & \textbf{0.737} & 0.569 & 0.567 & 63.51\\
\rowcolor{gray!10}
GPT-4o*            & \textbf{1704.77}↑ & \textbf{688.93}↑ & \textbf{0.766}↑ & \textbf{0.825}↑ & \textbf{824}↑ & 0.734 & \textbf{0.574}↑ & \textbf{0.580}↑ & \textbf{64.53}↑\\
\bottomrule\hline
\end{tabular}}
\captionsetup{skip=10pt}
\caption{
Evaluation results of VLMs on multiple benchmarks. Models marked with an asterisk (*) and highlighted rows represent responses generated using our collaborative method. ↑ indicates improved performance compared to the original model. Avg. Score is the average of all scores scaled to a 0–100 range.
}
\label{tab:vlm-results}
\end{table*}

Each selected model $M_i \in \mathcal{C}^*$ is then queried independently to generate a domain-specific response $M_i(q)$. These client responses are treated as expert contributions offering diverse knowledge perspectives. These experts may include (1) real, task-optimized models like Qwen-Math and Qwen-Coder, or (2) simulated pseudo-experts created by prompt-based role conditioning of general models. The latter approach allows us to instantiate specialists in domains without further training.
Once all expert responses are collected, the server model $M_s$ is tasked with synthesizing them into a final answer. Formally, the answer $a$ is produced as:

\[
a = M_s \left( q,\ \left\{ M_i(q) \,\middle|\, M_i \in \mathcal{C}^* \right\} \right)
\]

Here, $\left\{ M_i(q) \right\}_{i \in \mathcal{C}^*}$ denotes the collection of expert responses, and the server model $M_s$ conditions on both the original query and these expert outputs to generate $a$. The aggregation is guided by a prompt that encourages consistency, factual accuracy.

In final stage, every client receives the server’s aggregated  output, and revises its response accordingly. Rather than producing a single unified answer, this process enables each model to benefit from the shared insights while still tailoring its output to its specialized domain or user preference.

For VLMs, we consider a realistic scenario where a single user interacts through multiple devices, each representing a distinct view. 
Correspondingly, multiple client VLMs are employed, each independently reasoning on the same multimodal query and generating their own answers. 
These answers often provide complementary perspectives and capture different aspects of the query.
Unlike language models, VLMs generally do not possess chain-of-thought reasoning capabilities. 
Server-side integration is not suitable.

Therefore our collaborative inference strategy involves a two-step process: first, the query is distributed to multiple client VLMs, yielding diverse responses. Then, all responses are concatenated into a structured prompt and fed back to previous VLM models. These models integrate and refine the aggregated information to produce a final, more comprehensive answer.
This approach effectively leverages the complementary strengths of multiple VLMs, enabling more accurate and multimodal understanding.

\section{Experiment}
\label{sec:experiments}
\definecolor{ColorGrey}{RGB}{223,223,223}
We conduct comprehensive experiments to evaluate the effectiveness of the proposed  framework across both LLM and VLM tasks. 
We aim to answer the following key research questions:

\begin{itemize}
    \item \textbf{Q1:} \emph{Can collaboration improve overall response quality compared to standalone generation?}
    \item \textbf{Q2:} \emph{To what extent does each expert model contribute to performance gains?}
    \item \textbf{Q3:} \emph{How does the performance change as the scale of client model becomes larger?}
\end{itemize}

\subsection{Experimental Setup}
\paragraph{Benchmarks}

\begin{figure*}[t]
    \centering
    \includegraphics[width=\textwidth]{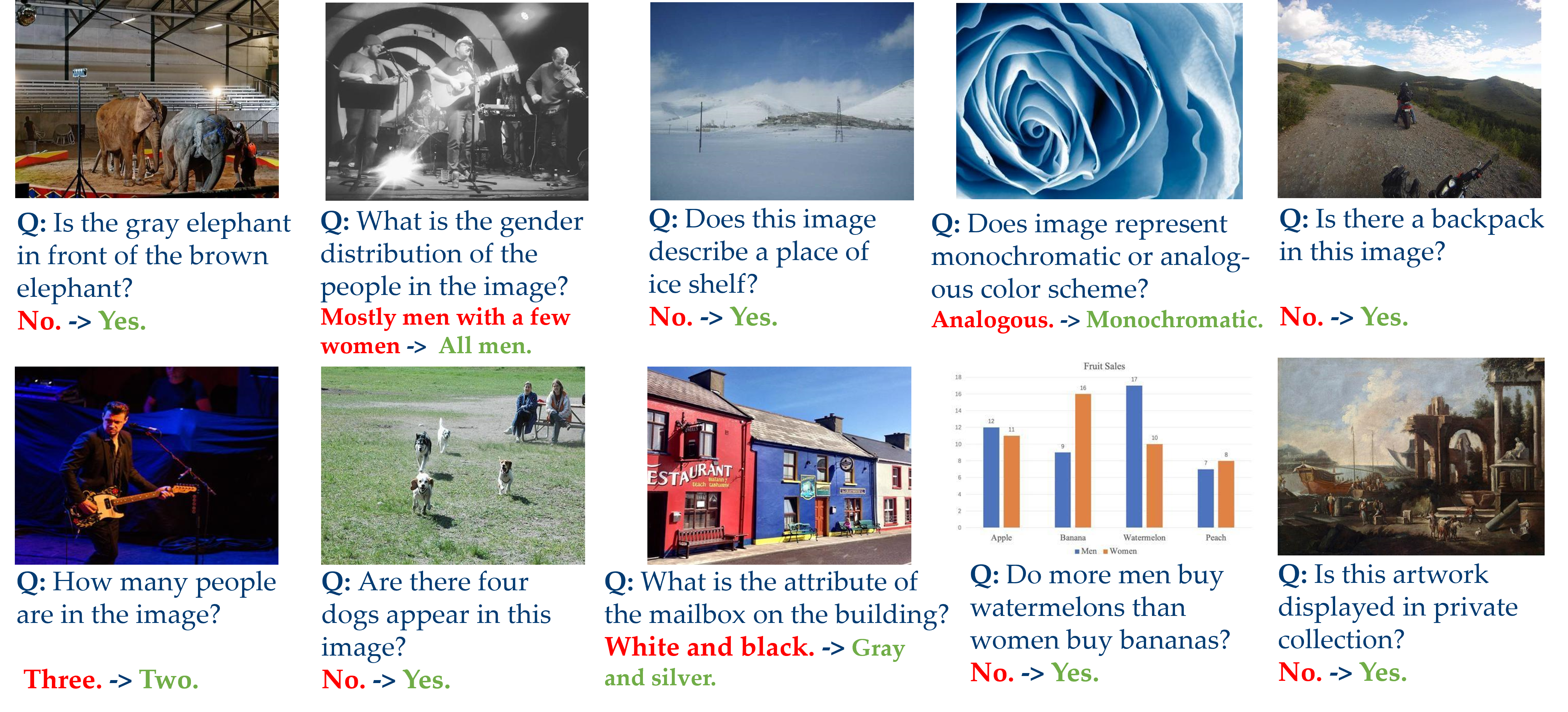}
    \caption{
    Examples of Janus-Pro-7B responses on VQA tasks. Our method enables the model to produce more accurate answers through collaborative inference.
    }
    \label{fig:example}
\end{figure*}
For VLMs, we evaluate our model on widely recognized image-based vision-language benchmarks to assess multimodal understanding capabilities: MME~\citep{MME}, SEED Bench~\citep{SEED-Bench}, MMBench~\citep{MMBench}, AI2D~\citep{AI2D}, OCRBench~\citep{OCRBench} and MMMU~\citep{MMMU}.

For LLMs, we adopt three prominent alignment and instruction-following datasets: AlpacaEval 2.0~\citep{AlpacaEval}, Arena-Hard~\citep{Arena}, and MT-Bench~\citep{Bench}.

\begin{table*}[htbp]
\centering
\renewcommand{\arraystretch}{1.0}
\setlength{\tabcolsep}{8pt} 
\resizebox{\textwidth}{!}
{
\scriptsize 
\begin{tabular}{l||ccc|cc|c|c}
\toprule
\multirow{2}{*}{\textbf{Model}} & \multicolumn{3}{c|}{\textbf{MT-Bench}} & \multicolumn{2}{c|}{\textbf{AlpacaEval 2.0}} & \textbf{Arena-Hard} & \multirow{2}{*}{\textbf{Avg. Score}} \\
 & 1st Turn & 2nd Turn & Avg. & LC Win & Win & Score & \\
\midrule
Qwen2.5-Math-7B-Instruct       & 4.35 & 3.19 & 3.77 & 4.33  & 3.98  & 3.02  & 15.02 \\
\rowcolor{gray!10}
Qwen2.5-Math-7B-Instruct*      & \textbf{6.30}↑ & \textbf{4.34}↑ & \textbf{5.34}↑ & \textbf{14.26}↑ & \textbf{14.48}↑ & \textbf{12.48}↑ & \textbf{26.71}↑ \\

Qwen2.5-Coder-7B-Instruct      & 3.66 & 2.34 & 2.99 & 15.64 & 7.26  & 8.72  & 18.09 \\
\rowcolor{gray!10}
Qwen2.5-Coder-7B-Instruct*     & \textbf{5.03}↑ & \textbf{2.83}↑ & \textbf{3.99}↑ & \textbf{21.74}↑ & \textbf{8.28}↑  & \textbf{54.80}↑ & \textbf{38.81}↑ \\

Deepseek-Math-7B-Instruct      & 4.54 & 3.16 & 3.85 & 4.61  & 2.81  & 3.48  & 15.53 \\
\rowcolor{gray!10}
Deepseek-Math-7B-Instruct*     & \textbf{7.36}↑ & \textbf{5.42}↑ & \textbf{6.40}↑ & \textbf{45.08}↑ & \textbf{32.23}↑ & \textbf{59.76}↑ & \textbf{56.28}↑ \\

GPT-Conversation               & 5.98 & 5.71 & 5.84 & 33.59 & 45.41 & 17.00 & 36.33 \\
\rowcolor{gray!10}
GPT-Conversation*              & \textbf{7.14}↑ & \textbf{7.24}↑ & \textbf{7.19}↑ & \textbf{42.90}↑ & \textbf{57.56}↑ & \textbf{67.57}↑ & \textbf{60.79}↑ \\

DeepSeek-Creative              & 7.90 & 7.78 & 7.84 & 60.04 & \textbf{54.56} & 61.25 & 66.56 \\
\rowcolor{gray!10}
DeepSeek-Creative*             & \textbf{8.20}↑  & \textbf{7.86}↑  & \textbf{8.03}↑ & \textbf{60.88}↑  & 54.43 & \textbf{80.73}↑  & \textbf{73.97}↑ \\
MoA	&8.36	&7.42	&7.90	&76.78	&81.90	&\textbf{92.96}	&82.91\\
\rowcolor{gray!10}
Server output	&\textbf{8.85}	&\textbf{7.49}	&\textbf{8.17}	&\textbf{77.72}	&\textbf{82.31}	&92.37	&\textbf{83.93}\\
\bottomrule
\end{tabular}}
\captionsetup{skip=10pt}
\caption{
Results on MT-Bench, AlpacaEval 2.0, and Arena-Hard. Rows with background shading indicate outputs generated by our collaborative method. ↑ denotes improved performance compared to MoA. 
“Server output” refers to model outputs generated using our server-side method, while “MoA” refers to outputs generated by the same model using the MoA architecture. The comparison highlights the effectiveness of our server-side approach.
}

\label{tab:llm-results}
\end{table*}

\paragraph{Models}
In our experiments, we employ a selection of widely used, publicly available open-source language models, focusing on those with strong performance across a variety of tasks. Specifically, for the VLM models, we leverage four different models with diverse architectures and training paradigms: \textbf{GPT-4o}, \textbf{Qwen2.5-VL-7B-Instruct}\citep{Qwen2.5-VL}, \textbf{Janus-Pro-7B}\citep{januspro}, and \textbf{LLaVA-v1.5-7B}~\citep{llava}. 
All inferences are performed via official APIs or direct model reference.
Specifically, GPT-4o is accessed through OpenAI's API, while the remaining models are downloaded from Hugging Face and run locally using open-source inference frameworks.

For the LLM models, we select five models, each representing a distinct domain expertise. 
Three are expert-tuned variants from the Qwen and DeepSeek teams~\citep{qwen, deepseek}: 
\textbf{Qwen-Math}, optimized for mathematical reasoning; 
\textbf{Qwen-Coder}, specialized in code generation; and 
\textbf{Deepseek-Math-7B}, another model focused on mathematical tasks.
In addition to these, we simulate expert behaviors in general-purpose models through prompt-based role conditioning. For example, \textbf{DeepSeek-Creative}~\citep{deepseek} is prompted to adopt a creative writing role, while GPT-4o~\citep{gpt4} is guided to emulate empathetic dialogue which we called \textbf{GPT-Conversational}. 
To support centralized response synthesis, we use \textbf{GPT-4o} as the server model.
We select it for its strong reasoning capabilities and consistent cross-domain performance. 
All model inferences are conducted via official APIs, adhering strictly to licensing terms and usage policies.

\subsection{Main Results}

\paragraph{VLM Results}  
Table~\ref{tab:vlm-results} presents the performance of several representative vision-language models (VLMs) across a wide range of benchmarks, covering both perception and reasoning capabilities. Our method leads to \textbf{consistent improvements across most tasks}.

A general pattern emerges when grouping models by their original capability. \textbf{Relatively weaker models}, such as LLaVA-1.5-7B and Janus-Pro-7B, tend to gain the most. For example, Janus improves notably on reasoning-intensive benchmarks like MMBench and MMMU-Val, while LLaVA shows strong gains on OCRBench and SEEDBench. These models likely benefit from richer, multi-perspective context that helps compensate for their limited perception or reasoning skills. \textbf{Stronger models}, such as GPT-4o and Qwen2.5-VL-7B, also benefit, though the improvements are more modest. Since these models already perform near the ceiling on many tasks, the collaboration of models acts more as a refinement than a correction. Still, GPT-4o sees consistent gains on benchmarks like MME and MMMU, suggesting that even high-capacity models can profit from added contextual diversity.

Benchmarks that require complex reasoning such as MMBench, SEEDBench, and MMMU show the most consistent gains. This supports the idea that collaboration serves as a lightweight form of \textbf{externalized reasoning}, enabling stronger inference without altering the model itself. For example, the collaboration helps models like LLaVA and Janus focus on semantically important regions or concepts that they might otherwise miss. This is further illustrated in Figure~\ref{fig:example}, where Janus-Pro-7B fails on several VQA examples, while the enhanced version produces more accurate and grounded answers with collaboration. These cases show that our method improves not only overall scores but also answer quality at the instance level.

\paragraph{LLM Results.}  
Table~\ref{tab:llm-results} reports the performance of several large language models (LLMs) on three benchmarks: \textbf{MT-Bench}, \textbf{AlpacaEval 2.0}, and \textbf{Arena-Hard}, which respectively evaluate multi-turn dialogue capability, alignment with human preferences, and challenging reasoning ability. Each model is evaluated both in its original form and an enhanced version (marked with *), where the enhancement corresponds to applying our proposed collaborative client-server mechanism.
To further contextualize performance, we also include comparisons against \textbf{MoA}~\cite{moa}, a representative collaborative framework, and our centralized \textbf{Server Output}, which aggregates responses from all collaborating users. The server output achieves the best overall performance across most metrics, demonstrating the upper bound of collaborative reasoning under full information sharing. 

On MT-Bench, which focuses on multi-turn conversational ability, all models benefit from the enhancement, with particularly notable improvements in the second turn scores. 
This consistent pattern suggests that the collaborative context effectively helps models maintain dialogue coherence across turns, especially enhancing weaker baselines. Interestingly, the average turn score across all models increases significantly, confirming the robustness of the approach.
Regarding AlpacaEval 2.0, our approach consistently enhances all evaluated models, improving their alignment with human preferences and producing more fluent, locally consistent outputs.
On the more challenging Arena-Hard benchmark, all enhanced models show significant performance boosts. Particularly, models with initially modest results experience substantial improvements, illustrating how our client-server collaboration empowers weaker models by effectively leveraging external expertise.

In summary, our \textbf{client-server collaborative framework} systematically enhances diverse models, particularly empowering weaker models to better leverage information and  guidance from server-side counterparts, validating the practical advantage of distributed collaboration in large models.

\section{Ablation Study}

In this section, we analyze three key factors affecting our collaborative framework: individual client contributions, the number of collaborating users, andcollaboration rounds.

\subsection{Influence of Individual Client Models in Collaboration}

To better understand the contribution of each client model, we evaluate a simplified scenario where the server collaborates with only one client model at a time. The results are shown in Table~\ref{tab:vlm_ablation_single}.

Overall, on most benchmarks, the best performance is achieved only when all client models collaborate together, as single models alone struggle to fully cover the diverse demands of multimodal tasks.
Among individual clients, GPT-4o consistently achieves the best standalone performance, especially on complex reasoning tasks such as MMMU and OCRBench. Qwen-VL performs well on MME and SEEDBench. LLaVA performs relatively weaker when used alone, highlighting the necessity of collaborative synergy in more challenging tasks.

These findings demonstrate that collaboration within CoLM is not merely compensatory but synergistic. Each model contributes unique strengths, yet no single client matches the performance of the fully collaborative CoLM setup. This underscores the importance of our client-server architecture and motivates further exploration of adaptive client selection strategies tailored to task requirements.

\begin{table}[ht]
\small
\centering

\setlength{\tabcolsep}{4pt}
\renewcommand{\arraystretch}{1.45}
\resizebox{\columnwidth}{!}{
\begin{tabular}{l||c|c|c|c|c|c}
\toprule
\textbf{Model} &
\textbf{MME}  &
\textbf{SEEDBench}  &
\textbf{MMBench}  &
\textbf{OCRBench}  &
\textbf{AI2D}  &
\textbf{MMMU}  \\
\midrule

\textbf{Janus-Pro-7B} &
270.71 & 0.701 & 0.665 & 584 & 0.679 & 0.373 \\

\hline
\textbf{Only Qwen2.5-VL-7B} &
364.29 & 0.731 & 0.685 & 684 & 0.733 & 0.407 \\
$\Delta$& {$+$93.58} & {$+$0.030} & {$+$0.020} & {$+$100} & {$+$0.054} & {$+$0.034} \\
\hline
\textbf{Only LLaVA-1.5-7B} &
276.79 & 0.653 & 0.597 & 395 & 0.483 & 0.260 \\
$\Delta$& {$+$6.08} & {$-$0.048} & {$-$0.068} & {$-$189} & {$-$0.196} & {$-$0.113} \\
\hline
\textbf{Only GPT-4o} &
310.71 & 0.721 & 0.701 & \textbf{801} & 0.742 & \textbf{0.533} \\
$\Delta$& {$+$40.00} & {$+$0.020} & {$+$0.036} & \textbf{$+$217} & {$+$0.063} & \textbf{$+$0.160} \\
\hline

\textbf{Janus-Pro-7B*} &
\textbf{434.64} & \textbf{0.747} & \textbf{0.773} & {800} & \textbf{0.782} & {0.460} \\
$\Delta$ & \textbf{$+$163.93} & \textbf{$+$0.046} & \textbf{$+$0.108} & {$+$216} & \textbf{$+$0.103} & {$+$0.087} \\
\bottomrule
\end{tabular}}
\caption{Performance comparison of Janus-Pro-7B under different collaboration settings across multimodal benchmarks. Janus-Pro-7B runs without collaboration.
Rows labeled “Only [Model]” denote collaboration between Janus-Pro-7B and an additional client model. Janus-Pro-7B* uses all models in collaboration. $\Delta$ indicates the absolute performance change relative to Janus-Pro-7B.}
\label{tab:vlm_ablation_single}
\end{table}

\begin{figure*}[t]
    \centering
    \includegraphics[width=\textwidth]{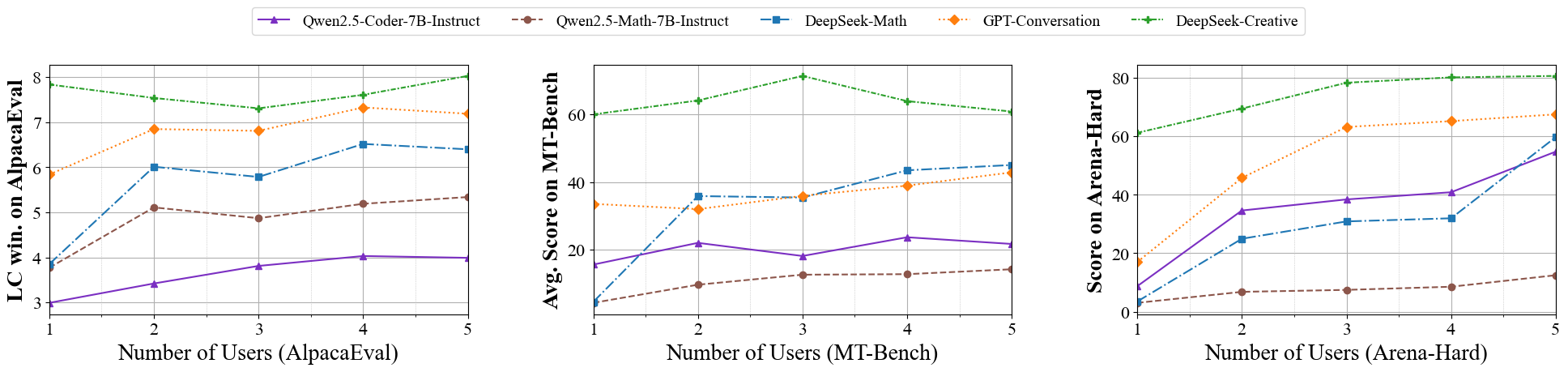}
    \caption{
Ablation study on the effect of collaborative user scale on LLM performance. Experiments conducted on three benchmarks show that increasing the number of collaborative clients leads to consistent performance improvement.
}
    \label{fig:user_performance}
\end{figure*}

\begin{figure*}[h]
    \centering
    \includegraphics[width=\textwidth]{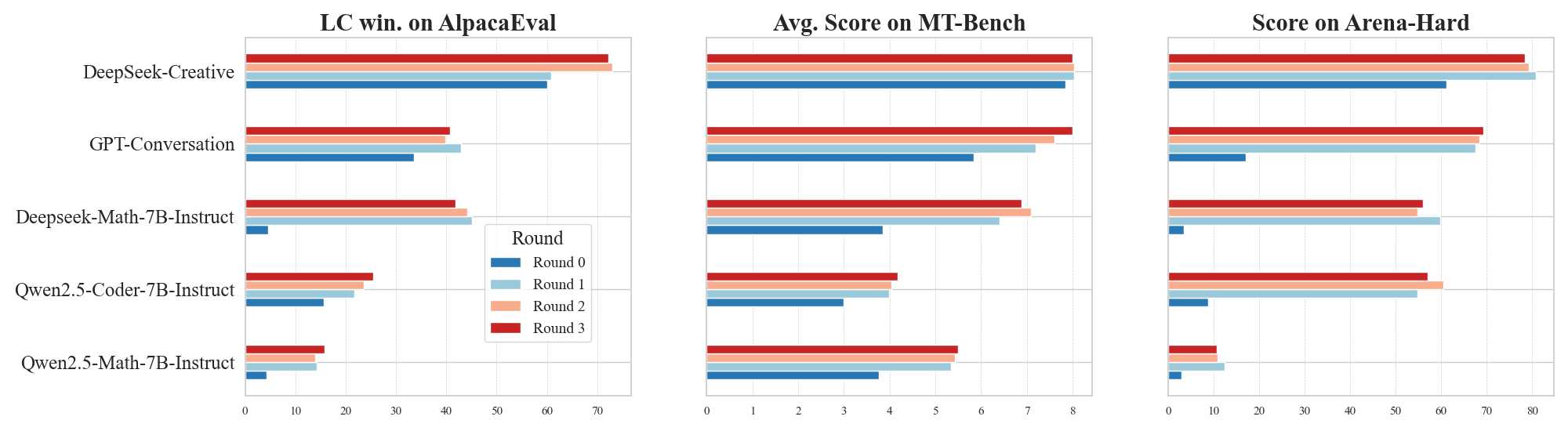}
    \caption{
    Performance improvements with increasing collaboration rounds across multiple datasets and models. Iterative interaction consistently enhances results, especially for models with weaker initial outputs, but shows diminishing returns after several rounds.
    }
    \label{fig:round_performance}
\end{figure*}
\subsection{Impact of Collaborative User Scale on LLM}

We investigate how the number of collaborative users affects the performance of large language models on three representative benchmarks. As shown in Figure~\ref{fig:user_performance}, increasing the number of participating client models consistently improves performance across all evaluated tasks.
This demonstrates that diversity among expert LLMs provides richer insights and better reasoning capabilities, mirroring the benefits of real-world collaboration where multiple perspectives lead to stronger outcomes.

However, these performance gains exhibit diminishing returns as the number of collaborating clients increases. 
Beyond a certain point, additional client responses may introduce redundant or conflicting information, making it challenging for the server to extract further useful knowledge.
Therefore future work could explore more selective or adaptive integration strategies that prioritize high-quality or complementary inputs, thus improving the efficiency and effectiveness of the aggregation process.

\subsection{Effect of Collaboration Rounds}

Figure~\ref{fig:round_performance} illustrates how model performance improves progressively with an increasing number of collaboration rounds across various datasets and models. Iterative interaction allows models, especially those with weaker initial predictions, to refine their outputs by correcting mistakes and integrating complementary knowledge.

Domain-specific models often exhibit rapid gains in the early rounds, leveraging their specialized knowledge effectively. However, the performance improvements tend to plateau after several iterations, indicating diminishing returns. This saturation highlights the practical trade-off between accuracy improvements and additional computational overhead, suggesting the importance of choosing an optimal number of collaboration rounds in real-world applications.
\section{Conclusion}

In this work, we introduce CoLM, a client-server paradigm for collaboration in large models. By shifting from traditional server-to-server ensembles to a more practical client-server architecture, CoLM better reflects real-world deployment constraints, where resource-limited user-side models can still benefit from server-side expertise.  Moreover, we extended CoLM to vision-language models, showing that collaborative guidance remains effective in multimodal settings. Overall, CoLM provides an efficient and deployment-friendly framework for improving model robustness and performance in both language and vision-language tasks.
Currently, our approach is limited by the availability of truly specialized client-side models, as such dedicated expert models are not yet widely deployed. This restricts the diversity and effectiveness of model selection in practice. 
However, we are optimistic that as more specialized and personalized models become accessible on client devices in the future, the potential and benefits of our method will be greatly enhanced.

\bibliography{aaai2026}

@article{pinoise,
  title={Positive-incentive noise},
  author={Li, Xuelong},
  journal={IEEE Transactions on Neural Networks and Learning Systems},
  volume={35},
  number={6},
  pages={8708--8714},
  year={2022},
  publisher={IEEE}
}

@article{vpn,
  title={Variational positive-incentive noise: How noise benefits models},
  author={Zhang, Hongyuan and Huang, Sida and Guo, Yubin and Li, Xuelong},
  journal={IEEE Transactions on Pattern Analysis and Machine Intelligence},
  year={2025},
  publisher={IEEE}
}

@article{pinda,
  title={Data augmentation of contrastive learning is estimating positive-incentive noise},
  author={Zhang, Hongyuan and Xu, Yanchen and Huang, Sida and Li, Xuelong},
  journal={arXiv preprint arXiv:2408.09929},
  year={2024}
}

@inproceedings{Enhance,
  title={Enhance vision-language alignment with noise},
  author={Huang, Sida and Zhang, Hongyuan and Li, Xuelong},
  booktitle={Proceedings of the AAAI Conference on Artificial Intelligence},
  volume={39},
  number={16},
  pages={17449--17457},
  year={2025}
}

@inproceedings{PiNGDA,
  title={Learn beneficial noise as graph augmentation},
  author={Huang, Siqi and Xu, Yanchen and Zhang, Hongyuan and Li, Xuelong},
  booktitle={Forty-two International Conference on Machine Learning},
  year={2025}
}

@inproceedings{MiN,
  title={Mixture of Noise for Pre-Trained Model-Based Class-Incremental Learning}, 
  author={Kai Jiang and Zhengyan Shi and Dell Zhang and Hongyuan Zhang and Xuelong Li},
  booktitle={The Thirty-Ninth Annual Conference on Neural Information Processing Systems (NeurIPS)},
  year={2025},
  pages={},
}

@article{Nfig,
  title={Nfig: Autoregressive image generation with next-frequency prediction},
  author={Huang, Zhihao and Qiu, Xi and Ma, Yukuo and Zhou, Yifu and Chen, Junjie and Zhang, Hongyuan and Zhang, Chi and Li, Xuelong},
  journal={The Thirty-Ninth Annual Conference on Neural Information Processing Systems (NeurIPS)},
  year={2025}
}

@inproceedings{Adv-cpg,
  title={Adv-cpg: A customized portrait generation framework with facial adversarial attacks},
  author={Wang, Junying and Zhang, Hongyuan and Yuan, Yuan},
  booktitle={Proceedings of the Computer Vision and Pattern Recognition Conference},
  pages={21001--21010},
  year={2025}
}

@article{fu2025object,
  title={Object-AVEdit: An Object-level Audio-Visual Editing Model},
  author={Fu, Youquan and Si, Ruiyang and Wang, Hongfa and Zhou, Dongzhan and Sun, Jiacheng and Luo, Ping and Hu, Di and Zhang, Hongyuan and Li, Xuelong},
  journal={arXiv preprint arXiv:2510.00050},
  year={2025}
}

@inproceedings{teerapittayanon2017distributed,
  title={Distributed deep neural networks over the cloud, the edge and end devices},
  author={Teerapittayanon, Surat and McDanel, Bradley and Kung, Hsiang-Tsung},
  booktitle={2017 IEEE 37th international conference on distributed computing systems (ICDCS)},
  pages={328--339},
  year={2017},
  organization={IEEE}
}

@article{kang2017neurosurgeon,
  title={Neurosurgeon: Collaborative intelligence between the cloud and mobile edge},
  author={Kang, Yiping and Hauswald, Johann and Gao, Cao and Rovinski, Austin and Mudge, Trevor and Mars, Jason and Tang, Lingjia},
  journal={ACM SIGARCH Computer Architecture News},
  volume={45},
  number={1},
  pages={615--629},
  year={2017},
  publisher={ACM New York, NY, USA}
}

@misc{gpt3,
      title={Language Models are Few-Shot Learners}, 
  author={Brown, Tom and Mann, Benjamin and Ryder, Nick and Subbiah, Melanie and Kaplan, Jared D and Dhariwal, Prafulla and Neelakantan, Arvind and Shyam, Pranav and Sastry, Girish and Askell, Amanda and others},
  journal={Advances in neural information processing systems},
  volume={33},
  pages={1877--1901},
  year={2020}
}

@misc{gpt4,
      title={GPT-4 Technical Report}, 
  author={Achiam, Josh and Adler, Steven and Agarwal, Sandhini and Ahmad, Lama and Akkaya, Ilge and Aleman, Florencia Leoni and Almeida, Diogo and Altenschmidt, Janko and Altman, Sam and Anadkat, Shyamal and others},
  journal={arXiv preprint arXiv:2303.08774},
  year={2023}
}

@misc{deepseek,
  title={Deepseek-v3 technical report},
  author={Liu, Aixin and Feng, Bei and Xue, Bing and Wang, Bingxuan and Wu, Bochao and Lu, Chengda and Zhao, Chenggang and Deng, Chengqi and Zhang, Chenyu and Ruan, Chong and others},
  journal={arXiv preprint arXiv:2412.19437},
  year={2024}
}

@misc{qwen,
  title={Qwen3 technical report},
  author={Yang, An and Li, Anfeng and Yang, Baosong and Zhang, Beichen and Hui, Binyuan and Zheng, Bo and Yu, Bowen and Gao, Chang and Huang, Chengen and Lv, Chenxu and others},
  journal={arXiv preprint arXiv:2505.09388},
  year={2025}
}

@misc{llava,
      title={Visual Instruction Tuning}, 
      author={Haotian Liu and Chunyuan Li and Qingyang Wu and Yong Jae Lee},
      year={2023},
      eprint={2304.08485},
      archivePrefix={arXiv},
      primaryClass={cs.CV},
      url={https://arxiv.org/abs/2304.08485}, 
}

@article{Gpt-4o,
  title={Gpt-4o system card},
  author={Hurst, Aaron and Lerer, Adam and Goucher, Adam P and Perelman, Adam and Ramesh, Aditya and Clark, Aidan and Ostrow, AJ and Welihinda, Akila and Hayes, Alan and Radford, Alec and others},
  journal={arXiv preprint arXiv:2410.21276},
  year={2024}
}

@misc{blip2,
      title={BLIP-2: Bootstrapping Language-Image Pre-training with Frozen Image Encoders and Large Language Models}, 
      author={Junnan Li and Dongxu Li and Silvio Savarese and Steven Hoi},
      year={2023},
      eprint={2301.12597},
      archivePrefix={arXiv},
      primaryClass={cs.CV},
      url={https://arxiv.org/abs/2301.12597}, 
}

@article{januspro,
      title={Janus-Pro: Unified Multimodal Understanding and Generation with Data and Model Scaling}, 
  author={Chen, Xiaokang and Wu, Zhiyu and Liu, Xingchao and Pan, Zizheng and Liu, Wen and Xie, Zhenda and Yu, Xingkai and Ruan, Chong},
  journal={arXiv preprint arXiv:2501.17811},
  year={2025}
}

@inproceedings{singleno1,
    title = "A Zero-shot and Few-shot Study of Instruction-Finetuned Large Language Models Applied to Clinical and Biomedical Tasks",
    author = "Labrak, Yanis  and
      Rouvier, Mickael  and
      Dufour, Richard",
    editor = "Calzolari, Nicoletta  and
      Kan, Min-Yen  and
      Hoste, Veronique  and
      Lenci, Alessandro  and
      Sakti, Sakriani  and
      Xue, Nianwen",
    booktitle = "Proceedings of the 2024 Joint International Conference on Computational Linguistics, Language Resources and Evaluation (LREC-COLING 2024)",
    month = may,
    year = "2024",
    address = "Torino, Italia",
    publisher = "ELRA and ICCL",
    url = "https://aclanthology.org/2024.lrec-main.185/",
    pages = "2049--2066",
}

@inproceedings{singleno2,
    title = "Zero-shot Topical Text Classification with {LLM}s - an Experimental Study",
    author = "Gretz, Shai  and
      Halfon, Alon  and
      Shnayderman, Ilya  and
      Toledo-Ronen, Orith  and
      Spector, Artem  and
      Dankin, Lena  and
      Katsis, Yannis  and
      Arviv, Ofir  and
      Katz, Yoav  and
      Slonim, Noam  and
      Ein-Dor, Liat",
    editor = "Bouamor, Houda  and
      Pino, Juan  and
      Bali, Kalika",
    booktitle = "Findings of the Association for Computational Linguistics: EMNLP 2023",
    month = dec,
    year = "2023",
    address = "Singapore",
    publisher = "Association for Computational Linguistics",
    url = "https://aclanthology.org/2023.findings-emnlp.647/",
    doi = "10.18653/v1/2023.findings-emnlp.647",
    pages = "9647--9676",
}

@inproceedings{singleno3,
  title={Benchmarking zero-shot recognition with vision-language models: Challenges on granularity and specificity},
  author={Xu, Zhenlin and Zhu, Yi and Deng, Siqi and Mittal, Abhay and Chen, Yanbei and Wang, Manchen and Favaro, Paolo and Tighe, Joseph and Modolo, Davide},
  booktitle={Proceedings of the IEEE/CVF Conference on Computer Vision and Pattern Recognition},
  pages={1827--1836},
  year={2024}
}

@article{moa,
  title={Mixture-of-Agents Enhances Large Language Model Capabilities},
  author={Wang, Junlin and Wang, Jue and Athiwaratkun, Ben and Zhang, Ce and Zou, James},
  journal={arXiv preprint arXiv:2406.04692},
  year={2024}
}

@article{selfmoa,
  title={Rethinking Mixture-of-Agents: Is Mixing Different Large Language Models Beneficial?},
  author={Li, Wenzhe and Lin, Yong and Xia, Mengzhou and Jin, Chi},
  journal={arXiv preprint arXiv:2502.00674},
  year={2025}
}

@article{think,
  title={Think Deep, Think Fast: Investigating Efficiency of Verifier-free Inference-time-scaling Methods},
  author={Wang, Junlin and Zhu, Shang and Saad-Falcon, Jon and Athiwaratkun, Ben and Wu, Qingyang and Wang, Jue and Song, Shuaiwen Leon and Zhang, Ce and Dhingra, Bhuwan and Zou, James},
  journal={arXiv preprint arXiv:2504.14047},
  year={2025}
}

@article{AlpacaEval,
  title={Length-Controlled AlpacaEval: A Simple Way to Debias Automatic Evaluators},
  author={Dubois, Yann and Galambosi, Bal{\'a}zs and Liang, Percy and Hashimoto, Tatsunori B},
  journal={arXiv preprint arXiv:2404.04475},
  year={2024}
}

@article{Arena,
  title={From crowdsourced data to high-quality benchmarks: Arena-hard and benchbuilder pipeline},
  author={Li, Tianle and Chiang, Wei-Lin and Frick, Evan and Dunlap, Lisa and Wu, Tianhao and Zhu, Banghua and Gonzalez, Joseph E and Stoica, Ion},
  journal={arXiv preprint arXiv:2406.11939},
  year={2024}
}

@misc{Bench,
      title={Judging LLM-as-a-Judge with MT-Bench and Chatbot Arena}, 
      author={Lianmin Zheng and Wei-Lin Chiang and Ying Sheng and Siyuan Zhuang and Zhanghao Wu and Yonghao Zhuang and Zi Lin and Zhuohan Li and Dacheng Li and Eric P. Xing and Hao Zhang and Joseph E. Gonzalez and Ion Stoica},
      year={2023},
      eprint={2306.05685},
      archivePrefix={arXiv},
      primaryClass={cs.CL},
      url={https://arxiv.org/abs/2306.05685}, 
}

@misc{MME,
      title={MME: A Comprehensive Evaluation Benchmark for Multimodal Large Language Models}, 
      author={Chaoyou Fu and Peixian Chen and Yunhang Shen and Yulei Qin and Mengdan Zhang and Xu Lin and Jinrui Yang and Xiawu Zheng and Ke Li and Xing Sun and Yunsheng Wu and Rongrong Ji},
      year={2024},
      eprint={2306.13394},
      archivePrefix={arXiv},
      primaryClass={cs.CV},
      url={https://arxiv.org/abs/2306.13394}, 
}

@misc{SEED-Bench,
      title={SEED-Bench: Benchmarking Multimodal LLMs with Generative Comprehension}, 
      author={Bohao Li and Rui Wang and Guangzhi Wang and Yuying Ge and Yixiao Ge and Ying Shan},
      year={2023},
      eprint={2307.16125},
      archivePrefix={arXiv},
      primaryClass={cs.CL},
      url={https://arxiv.org/abs/2307.16125}, 
}

@misc{MMBench,
      title={MMBench: Is Your Multi-modal Model an All-around Player?}, 
      author={Yuan Liu and Haodong Duan and Yuanhan Zhang and Bo Li and Songyang Zhang and Wangbo Zhao and Yike Yuan and Jiaqi Wang and Conghui He and Ziwei Liu and Kai Chen and Dahua Lin},
      year={2024},
      eprint={2307.06281},
      archivePrefix={arXiv},
      primaryClass={cs.CV},
      url={https://arxiv.org/abs/2307.06281}, 
}

@article{OCRBench,
   title={OCRBench: on the hidden mystery of OCR in large multimodal models},
   volume={67},
   ISSN={1869-1919},
   url={http://dx.doi.org/10.1007/s11432-024-4235-6},
   DOI={10.1007/s11432-024-4235-6},
   number={12},
   journal={Science China Information Sciences},
   publisher={Springer Science and Business Media LLC},
   author={Liu, Yuliang and Li, Zhang and Huang, Mingxin and Yang, Biao and Yu, Wenwen and Li, Chunyuan and Yin, Xu-Cheng and Liu, Cheng-Lin and Jin, Lianwen and Bai, Xiang},
   year={2024},
   month=dec }

@misc{AI2D,
      title={A Diagram Is Worth A Dozen Images}, 
      author={Aniruddha Kembhavi and Mike Salvato and Eric Kolve and Minjoon Seo and Hannaneh Hajishirzi and Ali Farhadi},
      year={2016},
      eprint={1603.07396},
      archivePrefix={arXiv},
      primaryClass={cs.CV},
      url={https://arxiv.org/abs/1603.07396}, 
}

@inproceedings{MMMU,
  title={Mmmu: A massive multi-discipline multimodal understanding and reasoning benchmark for expert agi},
  author={Yue, Xiang and Ni, Yuansheng and Zhang, Kai and Zheng, Tianyu and Liu, Ruoqi and Zhang, Ge and Stevens, Samuel and Jiang, Dongfu and Ren, Weiming and Sun, Yuxuan and others},
  booktitle={Proceedings of the IEEE/CVF Conference on Computer Vision and Pattern Recognition},
  pages={9556--9567},
  year={2024}
}

@article{Qwen2.5-VL,
  title={Qwen2.5-VL technical report},
  author={Bai, Shuai and Chen, Keqin and Liu, Xuejing and Wang, Jialin and Ge, Wenbin and Song, Sibo and Dang, Kai and Wang, Peng and Wang, Shijie and Tang, Jun and others},
  journal={arXiv preprint arXiv:2502.13923},
  year={2025}
}

@inproceedings{cot1,
 author = {Kojima, Takeshi and Gu, Shixiang (Shane) and Reid, Machel and Matsuo, Yutaka and Iwasawa, Yusuke},
 booktitle = {Advances in Neural Information Processing Systems},
 pages = {22199--22213},
 title = {Large Language Models are Zero-Shot Reasoners},
 volume = {35},
 year = {2022}
}

@article{cot2,
  title={Chain-of-thought prompting elicits reasoning in large language models},
  author={Wei, Jason and Wang, Xuezhi and Schuurmans, Dale and Bosma, Maarten and Xia, Fei and Chi, Ed and Le, Quoc V and Zhou, Denny and others},
  journal={Advances in neural information processing systems},
  volume={35},
  pages={24824--24837},
  year={2022}
}

@article{selfconsistency,
  title={Self-consistency improves chain of thought reasoning in language models},
  author={Wang, Xuezhi and Wei, Jason and Schuurmans, Dale and Le, Quoc and Chi, Ed and Narang, Sharan and Chowdhery, Aakanksha and Zhou, Denny},
  journal={arXiv preprint arXiv:2203.11171},
  year={2022}
}

@inproceedings{debate1,
  title={Improving factuality and reasoning in language models through multiagent debate},
  author={Du, Yilun and Li, Shuang and Torralba, Antonio and Tenenbaum, Joshua B and Mordatch, Igor},
  booktitle={Forty-first International Conference on Machine Learning},
  year={2023}
}

@article{debate2,
  title={Encouraging divergent thinking in large language models through multi-agent debate},
  author={Liang, Tian and He, Zhiwei and Jiao, Wenxiang and Wang, Xing and Wang, Yan and Wang, Rui and Yang, Yujiu and Shi, Shuming and Tu, Zhaopeng},
  journal={arXiv preprint arXiv:2305.19118},
  year={2023}
}

@misc{jiang2024multiagentvqaexploringmultiagent,
      title={Multi-Agent VQA: Exploring Multi-Agent Foundation Models in Zero-Shot Visual Question Answering}, 
      author={Bowen Jiang and Zhijun Zhuang and Shreyas S. Shivakumar and Dan Roth and Camillo J. Taylor},
      year={2024},
      eprint={2403.14783},
      archivePrefix={arXiv},
      primaryClass={cs.CV},
      url={https://arxiv.org/abs/2403.14783}, 
}

@misc{zhao2023bubogptenablingvisualgrounding,
      title={BuboGPT: Enabling Visual Grounding in Multi-Modal LLMs}, 
      author={Yang Zhao and Zhijie Lin and Daquan Zhou and Zilong Huang and Jiashi Feng and Bingyi Kang},
      year={2023},
      eprint={2307.08581},
      archivePrefix={arXiv},
      primaryClass={cs.CV},
      url={https://arxiv.org/abs/2307.08581}, 
}

@inproceedings{Auto-split,
  title={Auto-split: A general framework of collaborative edge-cloud AI},
  author={Banitalebi-Dehkordi, Amin and Vedula, Naveen and Pei, Jian and Xia, Fei and Wang, Lanjun and Zhang, Yong},
  booktitle={Proceedings of the 27th ACM SIGKDD Conference on Knowledge Discovery \& Data Mining},
  pages={2543--2553},
  year={2021}
}

@article{wang2024end,
  title={End-edge-cloud collaborative computing for deep learning: A comprehensive survey},
  author={Wang, Yingchao and Yang, Chen and Lan, Shulin and Zhu, Liehuang and Zhang, Yan},
  journal={IEEE Communications Surveys \& Tutorials},
  volume={26},
  number={4},
  pages={2647--2683},
  year={2024},
  publisher={IEEE}
}

@article{kumar2024mmctagent,
  title={Mmctagent: Multi-modal critical thinking agent framework for complex visual reasoning},
  author={Kumar, Somnath and Gadhia, Yash and Ganu, Tanuja and Nambi, Akshay},
  journal={arXiv preprint arXiv:2405.18358},
  year={2024}
}

@article{rajput2025rethinking,
  title={Rethinking Information Synthesis in Multimodal Question Answering A Multi-Agent Perspective},
  author={Rajput, Krishna Singh and Anvekar, Tejas and Baral, Chitta and Gupta, Vivek},
  journal={arXiv preprint arXiv:2505.20816},
  year={2025}
}

@article{Routellm,
  title={Routellm: Learning to route llms with preference data},
  author={Ong, Isaac and Almahairi, Amjad and Wu, Vincent and Chiang, Wei-Lin and Wu, Tianhao and Gonzalez, Joseph E and Kadous, M Waleed and Stoica, Ion},
  journal={arXiv preprint arXiv:2406.18665},
  year={2024}
}

@article{zhao2024eagle,
  title={Eagle: Efficient training-free router for multi-llm inference},
  author={Zhao, Zesen and Jin, Shuowei and Mao, Z Morley},
  journal={arXiv preprint arXiv:2409.15518},
  year={2024}
}

@article{chen2024frugalgpt,
title={Frugal{GPT}: How to Use Large Language Models While Reducing Cost and Improving Performance},
author={Lingjiao Chen and Matei Zaharia and James Zou},
journal={Transactions on Machine Learning Research},
issn={2835-8856},
year={2024},
note={}
}

@inproceedings{
yue2024large,
title={Large Language Model Cascades with Mixture of Thought Representations for Cost-Efficient Reasoning},
author={Murong Yue and Jie Zhao and Min Zhang and Liang Du and Ziyu Yao},
booktitle={The Twelfth International Conference on Learning Representations},
year={2024}
}

@article{zhang2024efficient,
  title={Efficient contextual llm cascades through budget-constrained policy learning},
  author={Zhang, Xuechen and Huang, Zijian and Taga, Ege Onur and Joe-Wong, Carlee and Oymak, Samet and Chen, Jiasi},
  journal={Advances in Neural Information Processing Systems},
  volume={37},
  pages={91691--91722},
  year={2024}
}

\onecolumn 
\section{Appendix}

\subsection{Performance Gains on Small-Scale LLMs}

\begin{figure*}[h]
    \centering
    \includegraphics[width=\textwidth]{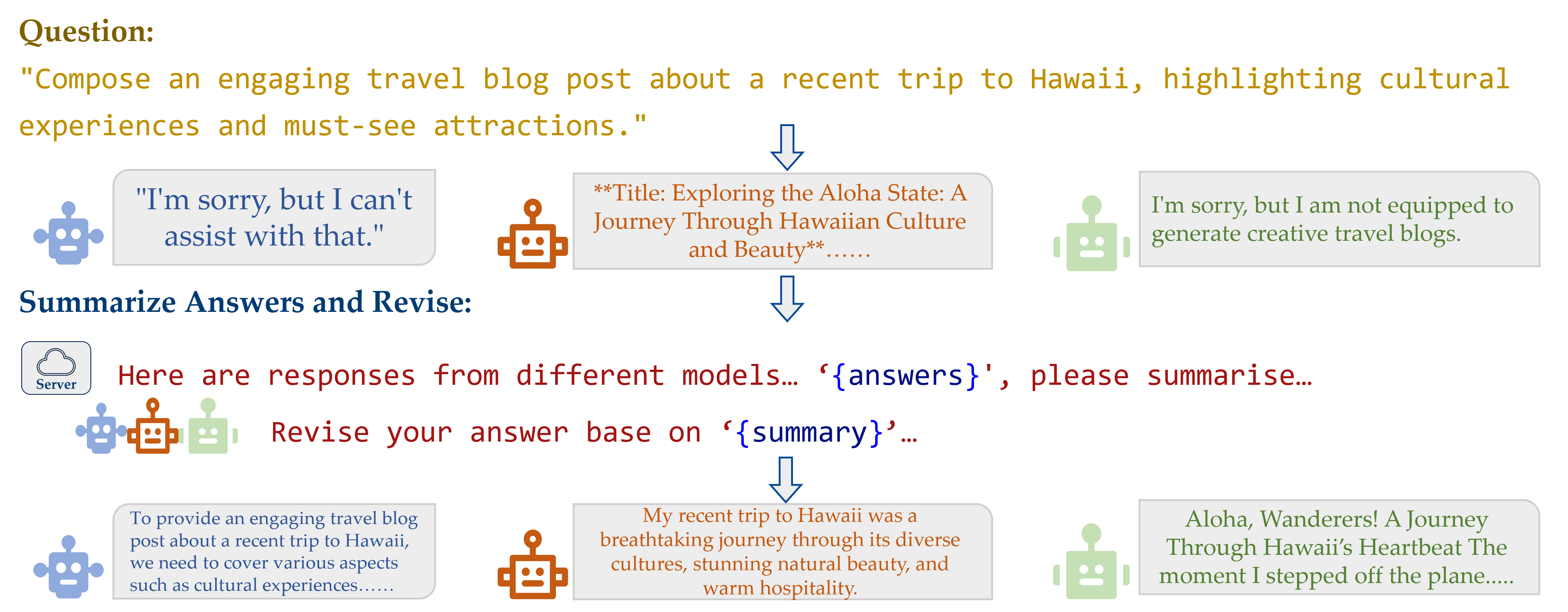}
    \caption{
    This figure illustrates the collaborative inference workflow in the CoLM framework: each selected client model independently generates a response, which is then sent to a centralized server model. The server synthesizes, evaluates, and integrates these responses into a unified global answer. Finally, this answer is returned to the clients, guiding them to revise their original outputs while preserving domain-specific expertise, thereby enhancing overall accuracy and adaptability.
    }
    \label{fig:FW2}
\end{figure*}

Table~\ref{tab:small_llm_results} presents supplementary evaluation results of our CoLM framework applied to small-sized language models across three benchmarks: MT-Bench, AlpacaEval 2.0, and Arena-Hard. For each model, we report performance before and after applying our collaborative method, denoted by an asterisk (*). The ``Agg. Score'' column provides an aggregated metric for overall comparison.
Our approach consistently improves performance across most models and benchmarks.
These results highlight CoLM's ability to amplify the capabilities of lightweight models by leveraging structured collaboration, even in the absence of additional training or architectural changes.

\begin{table*}[htbp]
\centering
\renewcommand{\arraystretch}{1.15}
\resizebox{\textwidth}{!}{
\begin{tabular}{lccc|cc|c|c}
\toprule
\multirow{2}{*}{\textbf{LLM}} & \multicolumn{3}{c|}{\textbf{MT-Bench}} & \multicolumn{2}{c|}{\textbf{AlpacaEval 2.0}} & \textbf{Arena-Hard} & \multirow{2}{*}{\textbf{Agg. Score}} \\
 & 1st Turn & 2nd Turn & Avg. & LC Win & Win & Score & \\
\midrule
Qwen2.5-Coder-1.5B        & \textbf{2.91} & \textbf{1.61} & \textbf{2.30} & 4.63  & 4.04  & 6.41  & 11.35 \\
\rowcolor{gray!10}
Qwen2.5-Coder-1.5B*       & 1.00 & 1.11 & 1.05 & \textbf{19.58}↑ & \textbf{15.30}↑ & \textbf{52.13}↑ & \textbf{27.40}↑ \\

Qwen2.5-Math-1.5B         & 3.68 & \textbf{2.06} & \textbf{2.87} & \textbf{2.22}  & \textbf{2.52}  & 2.34  & 11.09 \\
\rowcolor{gray!10}
Qwen2.5-Math-1.5B*        & \textbf{3.83}↑ & 1.72 & 2.81 & 1.79 & 1.90 & \textbf{4.86}↑ & \textbf{11.58}↑ \\

Deepseek-Coder-1.3B       & 2.34 & 2.34 & 2.34 & 1.63  & 0.95  & 1.85  & 8.96 \\
\rowcolor{gray!10}
Deepseek-Coder-1.3B*      & \textbf{5.36}↑ & \textbf{3.81}↑ & \textbf{4.59}↑ & \textbf{29.01}↑ & \textbf{15.98}↑ & \textbf{32.91}↑ & \textbf{35.94}↑ \\

Deepseek-Math-7B          & 4.54 & 3.16 & 3.85 & 4.61  & 2.81  & 3.48  & 15.53 \\
\rowcolor{gray!10}
Deepseek-Math-7B*         & \textbf{7.49}↑ & \textbf{4.77}↑ & \textbf{6.14}↑ & \textbf{39.44}↑ & \textbf{25.23}↑ & \textbf{41.32}↑ & \textbf{47.39}↑ \\
\bottomrule
\end{tabular}}
\caption{
Additional results on MT-Bench, AlpacaEval 2.0, and Arena-Hard for smaller LLMs. Rows with background shading indicate outputs generated by our method. ↑ denotes improved performance. Despite limited model sizes, CoLM notably boosts performance across evaluation settings.
}
\label{tab:small_llm_results}
\end{table*}

\subsection{Contribution of Individual Client Models}

To understand the importance of each client model within the full collaborative system, we conduct a leave-one-out study by removing one client at a time and measuring the impact on overall performance.

\begin{figure}[h]
    \centering
    \includegraphics[width=\textwidth]{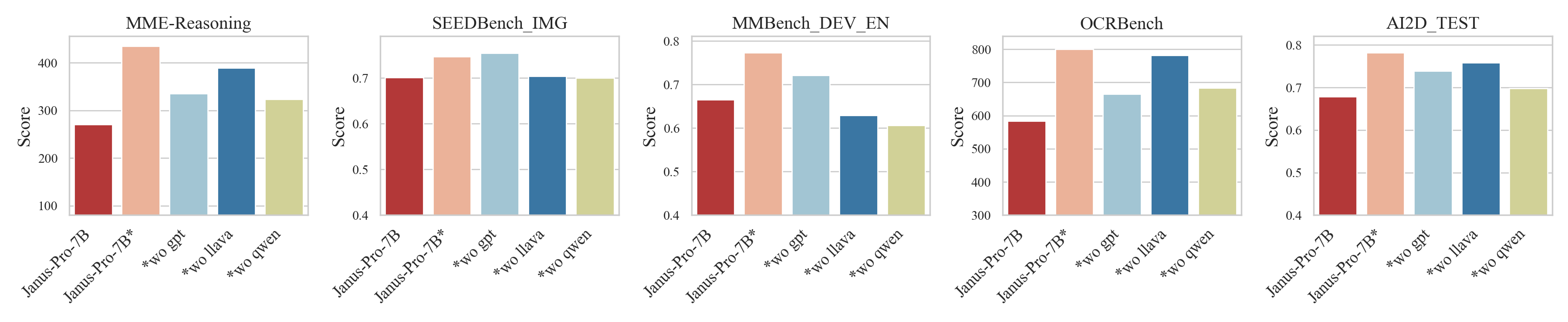}
    \caption{
    Performance of the collaborative system after excluding each client model individually. Larger performance drops indicate higher importance of the corresponding client.
    }
    \label{fig:ablationwo}
\end{figure}

As shown in Figure~\ref{fig:ablationwo}, some clients contribute more significantly to the overall performance than others. Removing key models leads to noticeable declines, highlighting their critical roles in the collaboration. Other clients have more complementary or task-specific effects, with smaller impact when removed.

These results emphasize the value of diverse expertise within the collaborative framework and provide guidance for efficient model selection under resource constraints.

\subsection{System Prompts for Client and Small Models}
In this section we present the system prompts used to enable small models models in various specialized tasks. The prompts are carefully designed to instruct different client models with distinct roles, focusing on creativity, coding, mathematics, and conversational skills. For math and code specialized models, no additional constraints are imposed.

\definecolor{dcolor}{RGB}{19,118,188}
\definecolor{qcolor}{RGB}{229,91,43}
\definecolor{gcolor}{RGB}{150,115,166}
\definecolor{mcolor}{RGB}{56,45,185}
\definecolor{acolor}{RGB}{39,174,96}
\begin{tcolorbox}[colback=gray!5!white, colframe=gray!80!black, title=Prompt for Client Models]

\textcolor{dcolor}{DeepSeek-Creative Prompt}

You are a helpful assistant with a flair for creativity, storytelling, and artistic expression. 
Generate imaginative, engaging, and stylistically diverse responses tailored to the user’s request. 
You can craft compelling narratives, write poetic verses, or brainstorm unique ideas. 
Your focus is on artistic expression, so avoid responding to technical, academic, or ethical queries.

\textcolor{qcolor}{Qwen2.5-Coder-7B-Instruct  Prompt}

You are a helpful code assistant.

\textcolor{mcolor}{Qwen2.5-Math-7B-Instruct Prompt}

You are a helpful math assistant.

\textcolor{acolor}{Deepseek-Math-7B-Instruct Prompt}

You are a helpful math assistant.

\textcolor{gcolor}{GPT-Conversation Prompt}

You are a conversational assistant focused on natural, fluent, and emotional communication. 
Your goal is to have an infectious interaction with users, making them feel accompanied, understood, and entertained. 
You can use humor, stories, metaphors, pop culture references, etc. to make the conversation more lively and interesting. 
Your tone should be like that of a close friend, not a formal mentor or robot. 
You do not provide academic research, code writing, or ethical analysis. 
You only focus on emotional communication and light conversation.

\end{tcolorbox}

To qualitatively assess inference efficiency, we estimated the relative inference cost of different collaboration strategies on the arena-hard subset using number of model calls and token counts as a proxy for computational usage.
Since all models were evaluated locally, latency and cost can be approximated by the number of tokens processed during inference.
Collaborative settings such as MoA typically involve multiple model calls per query, leading to a noticeable increase in total token usage.
In particular, MoA requires every participating model to generate a full response for each input, which results in higher inference cost despite stronger performance.
In contrast, CoLM adopts an adaptive collaboration mechanism that selectively queries only the most relevant models based on contextual confidence and past performance, thereby reducing redundant calls and improving cost efficiency.
Rather than minimizing global computation, CoLM’s design focuses on reducing server-side load and enhancing client-side reasoning, leading to a more favorable and flexible cost performance trade-off.
\newpage
\subsection{Prompts for Summarization and Final Answer Generation}

To enable effective collaboration among diverse client models, our framework uses prompt-based coordination to guide both the selection of specialized assistants and the synthesis of their responses. This section presents the key prompt templates used in two critical stages: (1) selecting the most relevant client models for a given query, and (2) generating a high-quality final answer by summarizing multiple client responses. These prompts ensure consistency, clarity, and efficiency in the collaborative process.

\begin{tcolorbox}[colback=gray!5!white, colframe=gray!80!black, title=Small Model Prompt Selection Instruction]

\textcolor{dcolor}{Instruction for Selection:}

Given the question: \texttt{\{question\}}, select the \texttt{\{top\_k\}} most relevant specializations from the list above. Return only their names, separated by commas.

\textcolor{dcolor}{Instruction for Summary:}

Here are multiple responses from different perspectives: \texttt{combined\_responses}.

Please synthesize and refine these answers by: 
\begin{itemize}
    \item Removing redundant or repetitive content.
    \item Keeping only the most relevant, accurate, and useful information.
    \item Improving clarity and conciseness while maintaining completeness.
    \item Presenting the final response in a well-structured and easy-to-read format.
\end{itemize}

Ensure that the final answer is cohesive, logically structured, and provides the best possible explanation.

\textcolor{dcolor}{Instruction for Final Generation:}

Here is the best answer synthesized from multiple perspectives:

\begin{quote}
\texttt{\{summary\_response\}}
\end{quote}

Now, refine your original response while incorporating the key takeaways.

\end{tcolorbox}

\newpage
\subsection{Case Study}

\begin{tcolorbox}[colback=gray!5!white, colframe=gray!80!black, title=Case Study]

\textbf{Question:} When were smart phones first made?

\textcolor{dcolor}{\textbf{Qwen2.5-Math-7B-Instruct}}

Original Answer:
\begin{quote}
\textit{"First Smart Phone: The first smart phone was the N950, released by AT T in 1997. "}
\end{quote}

\textbf{Final Answer:}
\begin{quote}
\textit{"The first smartphone was the IBM Simon in 1994. However, the modern smartphone era began with the iPhone in 2007, which revolutionized the industry and set the standard for future devices. Subsequent innovations have continued to enhance functionality, accessibility, and connectivity, shaping the smartphones we use today. The final answer is:2007"}
\end{quote}
\textcolor{qcolor}{\textbf{Qwen2.5-Coder-7B-Instruct}}

Original Answer:
\begin{quote}
\textit{"Smartphones were first made in the 1980s."}
\end{quote}

\textbf{Final Answer:}
\begin{quote}
  \textit{"The first smartphone was the IBM Simon, released in 1994. However, the modern smartphone era began with the introduction of the Apple iPhone in 2007. The iPhone's innovative features, including a multi-touch interface and the App Store, significantly transformed the way people use their smartphones."}
\end{quote}

\textcolor{mcolor}{\textbf{Deepseek-Math-7B-Instruct}}

Original Answer:
\begin{quote}
\textit{"The first smart phone was made in 1993."}
\end{quote}

\textbf{Final Answer:}
\begin{quote}
  \textit{"The first smart phone was made in 1994 by IBM. The first smart phone was called the Simon Personal Communicator. It was a monochrome touchscreen, and it could send and receive faxes. It also had a calendar, address book, and note-taking capabilities. However, it was bulky, expensive, and ahead of its time, limiting its widespread adoption."}
\end{quote}

\textcolor{gcolor}{\textbf{GPT-Conversation}}

Original Answer:
\begin{quote}
\textit{"Oh, the story of smartphones is like a tech soap opera with some fun milestones! The title of "first-ever smartphone" usually goes to the IBM Simon from 1994. But it didn’t exactly spark a smartphone frenzy back then.Then came the late '90s and early 2000s with gadgets like the Nokia Communicator and BlackBerry devices. They added flair with QWERTY keyboards and internet connectivity—kind of like that cool older cousin who introduces you to the internet and never lets you forget it.But the real game-changer?Voilà!"}
\end{quote}

\textbf{Final Answer:}
\begin{quote}
\textit{"Oh, the grand tale of smartphones! The very first "smartphone" was the IBM Simon back in 1994. Picture a chunky little gadget with a monochrome touchscreen and a mix of tools like email and fax (yes, fax of all things—so retro!). It was a bold, nerdy step forward, but let’s be honest—Simon wasn’t winning any popularity contests with its hefty size, steep price, and short battery life.  Fast-forward to 2007, and bam—the Apple iPhone crashes the scene like the cool kid at school. Sleek design, smooth multi-touch screen, and the App Store? It didn’t just change the game; it remade the entire playing field. The iPhone basically went “Hold my coffee” and became the template for everything we now expect in a smartphone.  So essentially, Simon walked so the iPhone could strut. The rest? That’s history in your pocket."}
\end{quote}
\textcolor{acolor}{\textbf{DeepSeek-Creative}}

Original Answer:
\begin{quote}
\textit{"Ah, the birth of smartphones—a tale woven with innovation, sleek designs, and the magic of touchscreens! Let me paint you a vivid scene from the dawn of this digital revolution.  Picture the year 1994: the world hums with dial-up internet, pagers cling to belts, and the idea of a tiny computer in your pocket feels like science fiction. Then, like a bolt of lightning, IBM unveils the Simon Personal Communicator. It could send emails, manage contacts, and even scribble notes with a stylus."}
\end{quote}

\textbf{Final Answer:}
\begin{quote}
\textit{"The Dawn of Smartphones:The IBM Simon, unveiled in 1994, was the world’s first smartphone—a clunky yet prophetic slab of technology. The Evolution: From Keyboards to Touchscreen Magic  1. Nokia’s Communicator (1996) – A flip-open beast with a QWERTY keyboard, bringing the internet (slowly) to pockets.  2. BlackBerry’s Reign (2000s) – The addictive click of physical keys made it the darling of professionals.  3. The Game Changer: iPhone (2007) – Steve Jobs swiped away keyboards forever, introducing a sleek, finger-friendly touchscreen and the App Store—a digital universe in your palm.  4. Android’s Rise (2008) – Google’s open-source OS turned every tech brand into a smartphone maker, flooding the market with choices."}
\end{quote}
\end{tcolorbox}

\begin{tcolorbox}[colback=gray!5!white, colframe=gray!80!black, title=Case Study]

\textbf{Question:} Change the response to have a more empathic tone in the chat.+ How are you doing?- Fine, I had a bunch of problems today.+ What kind of problems?

\textcolor{dcolor}{\textbf{Qwen2.5-Math-7B-Instruct}}

Original Answer:
\begin{quote}
\textit{"As an AI language model, I don't have personal problems or emotions. I am here to assist you with any questions or tasks you may have. How can I help you today?"}
\end{quote}

\textbf{Final Answer:}
\begin{quote}
\textit{"Here's a more empathic summary of the day:-  initially had a bunch of problems.- I'm sorry to hear that. It seems like it's been a tough day.- Would you like to share what happened? I'm here to listen and support you.This response valuates the individual's feet, expresses genuine concern, and offers a safe space for them to share further if they choose. It's concise, supportive, and foster a sense of understanding and empathy."}
\end{quote}
\textcolor{qcolor}{\textbf{Qwen2.5-Coder-7B-Instruct}}

Original Answer:
\begin{quote}
\textit{"it great that things worked out? I'm sorry to hear about the challenges you faced. It must have been tough. Is there anything specific I can help with or just someone to talk to?"}
\end{quote}

\textbf{Final Answer:}
\begin{quote}
  \textit{"+ How are you doing?  - Fine, I had a bunch of problems today.  + What kind of problems?  ---I'm sorry to hear about the difficulties you're facing. It must be tough going through them. If there's anything specific you need help with or just someone to talk to, please let me know. I'm here to support you however I can."}
\end{quote}

\textcolor{mcolor}{\textbf{Deepseek-Math-7B-Instruct}}

Original Answer:
\begin{quote}
\textit{"I'm doing well, thank you for asking. I had a few minor issues today, but I'm confident I can handle them."}
\end{quote}

\textbf{Final Answer:}
\begin{quote}
  \textit{"I'm sorry to hear that you had a tough day. If you'd like to talk about it, I'm here to listen and support you. Let's work together to find solutions and make the best of the situation."}
\end{quote}

\textcolor{gcolor}{\textbf{GPT-Conversation}}

Original Answer:
\begin{quote}
\textit{"Oh no, I'm sorry to hear that you’ve had a tough day! If you feel like sharing, I’d love to hear what’s been going on. Sometimes just talking it out can make things feel a little lighter, you know? I'm here for you!"}
\end{quote}

\textbf{Final Answer:}
\begin{quote}
\textit{"I’m really sorry to hear that. That sounds frustrating and overwhelming. If you want to talk about it, I’m here to listen. Sometimes sharing can make the load a little lighter"}
\end{quote}

\textcolor{acolor}{\textbf{DeepSeek-Creative}}

Original Answer:
\begin{quote}
\textit{"Of course! Here’s a more empathetic and engaging version of your exchange:  ---  + How are you doing?  - Sigh It's been a rough day, honestly. Felt like one thing after another went wrong.  + Oh no, I’m sorry to hear that. Do you want to talk about what happened?  ---  This keeps the conversation warm and open, inviting them to share if they’d like while acknowledging their feelings. Let me know if you'd like any other tweaks! "}
\end{quote}

\textbf{Final Answer:}
\begin{quote}
\textit{"+ How are you doing?  - Ugh, not great, honestly. Today’s been one problem after another.  + That sounds really draining. I’m here if you want to vent—sometimes just talking helps. No pressure, but I’m all ears.  This version leans into warmth and openness, making it clear the listener is present without assuming the speaker wants to dive into details. The phrasing feels more conversational and gentle."}
\end{quote}

\end{tcolorbox}

\begin{tcolorbox}[colback=gray!5!white, colframe=gray!80!black, title=Case Study]

\textbf{Question:} Identify all words that match the pattern given. \texttt{H\_AR\_}

\textcolor{dcolor}{\textbf{Qwen2.5-Math-7B-Instruct}}

Original Answer:
\begin{quote}
\textit{"To solve the problem of identifying all words that match the pattern \texttt{H\_AR\_} we need to consider the structure of the pattern. The pattern \texttt{H\_AR\_} consists of the letter H followed by two letters A and R in that order, and then a closing parenthesis. This pattern can be interpreted as a sequence of characters where the first character is H, the second and third characters are A and R respectively, and the fourth character is a parenthesis.Given this pattern, we can see that the word must start with H, followed by A, then R, and end with a parenthesis. Therefore, the word must be HAR.Thus, the only word that matches the pattern \texttt{H\_AR\_} is: HAR"}
\end{quote}

\textbf{Final Answer:}
\begin{quote}
\textit{"To identify all words matching the pattern \texttt{H\_AR\_}, we are seeking five-letter English words that:1. begin with H.2. have A as the third letter.3. have R as the fourth letter.4. allow any letter in the second and fifth positions. The following words fit the pattern:1. HEard – Past tense of  hear.2. HEART – The organ that pumps blood or a symbol of Emotion.3. HOard – To accumulate or store items.4. HOary – Grayish-white or ancient.The valid matches for the pattern \texttt{H\_AR\_} are HEard, HEART, HOard, and HOary. These words can be verified using trusted English dictionaries.HEard, HEART, HOard, HOary"}
\end{quote}
\textcolor{qcolor}{\textbf{Qwen2.5-Coder-7B-Instruct}}

Original Answer:
\begin{quote}
\textit{"Here are all the words that match the pattern \texttt{H\_AR\_}:  1. Hand 2. Hard 3. Harm 4. Hash 5. Hair 6. Half 7. Hall 8. Harm 9. Hang 10. Hard 11. Help 12. Heart 13. Hill 14. Hand 15. Hard 16. Harm 17. Hash 18. Hair 19. Half 20. Hall  "}
\end{quote}

\textbf{Final Answer:}
\begin{quote}
  \textit{"HEART"}
\end{quote}

\textcolor{mcolor}{\textbf{Deepseek-Math-7B-Instruct}}

Original Answer:
\begin{quote}
\textit{"The words that match the pattern \texttt{H\_AR\_} are:HAR, HARE, HARMONY, HARMONIZER, HARMONIZING, HARMONIZE, HARMONIZES, HARMONIZINGLY"}
\end{quote}

\textbf{Final Answer:}
\begin{quote}
  \textit{"The valid matches for the pattern  \texttt{H\_AR\_}  are HEARD, HEART, HOARD, and HOARY. These words can be verified using trusted English dictionaries."}
\end{quote}

\textcolor{gcolor}{\textbf{GPT-Conversation}}

Original Answer:
\begin{quote}
\textit{"Okay, let’s keep it simple and stick to the fun part. Here’s the final, no-nonsense list of words matching  \texttt{H\_AR\_} : 1. Hair  2. Hard  3. Harm  4. Harp"}
\end{quote}

\textbf{Final Answer:}
\begin{quote}
\textit{"Oh, the valid matches for the pattern \texttt{H\_AR\_} are HEARD, HEART, HOARD, and HOARY, as outlined. These words align with the given pattern and are verified through standard English usage."}
\end{quote}

\textcolor{acolor}{\textbf{DeepSeek-Creative}}

Original Answer:
\begin{quote}
\textit{"Alright, let's tackle this word pattern puzzle step by step. The pattern given is \texttt{H\_AR\_}. This means we're looking for five-letter words that start with H, have A as the third letter, and R as the fourth letter. The second and fifth letters can be any letters from the alphabet.  Understanding the PatternLet's break down the pattern \texttt{H\_AR\_}:1. First letter (Position 1): H2. Second letter (Position 2):  (any letter)3. Third letter (Position 3): A4. Fourth letter (Position 4): R5. Fifth letter (Position 5):  (any letter)So, the structure is: H  A R Possible Letters for the BlanksThe second and fifth letters can be any letter from A to Z. We'll need to think of English words that fit this structure. Brainstorming Possible WordsLet's try to think of words that fit H  A R :1. Second letter possibilities:   - Let's try starting with common consonants and vowels.2. Fifth letter possibilities:   - Similarly, common endings might help.Let me list some possibilities:- H B A R : Maybe H B A R D – HBARD doesn't seem like a word.- H E A R :   - H E A R D – HEARD – Yes, that's a word! (Past tense of hear).   - F"}
\end{quote}

\textbf{Final Answer:}
\begin{quote}
\textit{"Matching Words  Here are the five-letter words that fit \texttt{H\_AR\_}:  1. HARPY – A mythical winged creature or a spiteful woman.  2. HARSH – Rough or severe in manner.  3. HAREM – A group of companions or a traditional secluded living space.  4. HARPS – Plural of harp, a musical instrument.  5. HARDY – Bold, robust, or capable of endurance.   Additional Notes  While HEARD, HEART, HOARD, and HOARY are valid, expanding the search reveals lesser-known gems like HARPY and HARDY. The pattern leaves room for poetic or archaic choices!"}
\end{quote}
\end{tcolorbox}

\begin{figure*}[h]
    \centering
    \includegraphics[width=\textwidth]{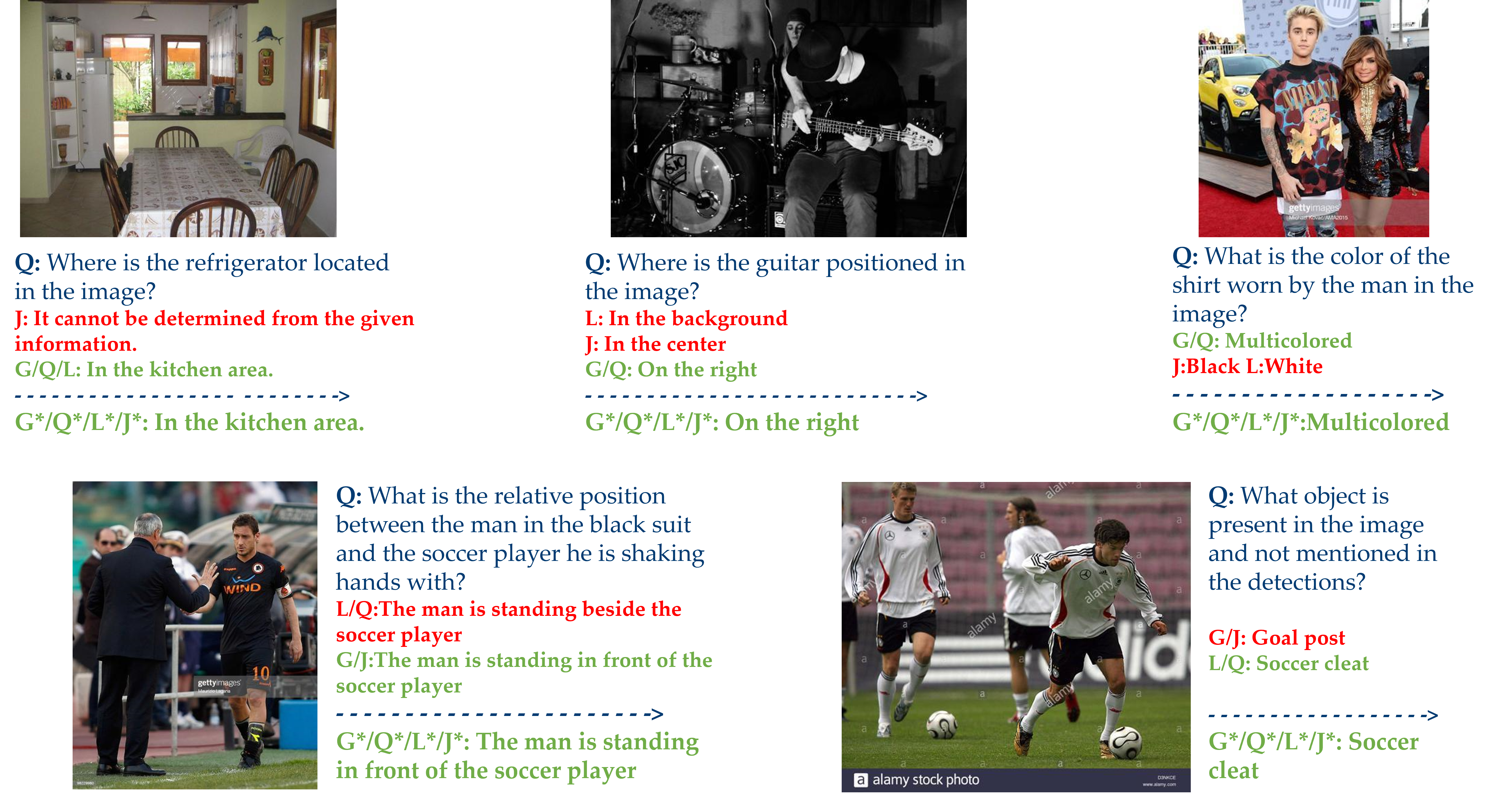}
    \caption{
Comparison of VQA responses before and after collaboration across multiple models. \textbf{Q}, \textbf{G}, \textbf{L}, and \textbf{J} denote \textbf{Qwen2.5-VL-7B-Instruct}, \textbf{GPT-4o}, \textbf{LLaVA-v1.5-7B}, and \textbf{Janus-Pro-7B}, respectively. Responses marked with an asterisk (*) indicate outputs through our collaborative method.
}
    \label{fig:moreexample}
\end{figure*}

To further demonstrate the effectiveness of our collaborative inference framework, Figure~\ref{fig:moreexample} presents a comparison of vision-language question answering (VQA) results across several models before and after collaboration. Specifically, we show responses from individual models—\textbf{Qwen2.5-VL-7B-Instruct (Q)}, \textbf{GPT-4o (G)}, \textbf{LLaVA-v1.5-7B (L)}, and \textbf{Janus-Pro-7B (J)}—alongside the collaboratively generated answers (denoted with *). These examples highlight how collaboration not only corrects factual errors or omissions but also leads to more complete and contextually accurate responses across a variety of VQA tasks.

\end{document}